\RequirePackage{amsmath}

\documentclass[
  a4paper,
  fontsize=10pt,
  parskip=half,
  oneside,
]{kaobook}

\usepackage[T1]{fontenc}
\usepackage[utf8]{inputenc}

\usepackage[scaled=0.92]{helvet}

\usepackage{xcolor}
\definecolor{burgundy}{HTML}{7B2D3A}
\definecolor{burgunlight}{HTML}{F5EEF0}
\definecolor{margnote}{HTML}{5A1F28}
\definecolor{ref_orange}{HTML}{fc911e}
\definecolor{link_red}{HTML}{99434d}

\usepackage{hyperref}
\hypersetup{
    colorlinks=true,
    urlcolor=link_red,
    linkcolor=link_red
}
\addtokomafont{chapter}{\color{burgundy}}
\addtokomafont{section}{\color{burgundy}}
\addtokomafont{subsection}{\color{burgundy}}

\usepackage[most]{tcolorbox}

\usepackage{booktabs}
\usepackage{enumitem}
\setlist[enumerate]{nosep, leftmargin=1.4em}
\setlist[itemize]{nosep, leftmargin=1.4em}

\usepackage{geometry}
\newcommand{\margintag}[2]{%
  \marginnote{\footnotesize\textbf{\color{burgundy}#1}\enspace #2}}

\usepackage{caption}

\usepackage[
  backend=biber,
  style=numeric,
  sorting=none,
  natbib=true,
  maxbibnames=10,
  minbibnames=10,
]{biblatex}
\providecommand{\addto}[2]{}
\usepackage{kaorefs}

\renewcommand{\thesection}{\arabic{section}}
\renewcommand{\thesubsection}{\arabic{section}.\arabic{subsection}}

\begin{document}
\pagestyle{centeredpagenum.scrheadings}

{\color{burgundy}\rule{\textwidth}{2pt}}
\vspace{0.5em}
{\fontsize{16}{20}\selectfont\bfseries\color{burgundy}
Agent Team Work Zone: \\An Automated, Persistent Workspace for \\ Long-Lived Claude Code Agent Teams\par}
\vspace{0.3em}
{\large Shouren~Wang\par}
\vspace{0.15em}
{\small Case Western Reserve University\par}
\vspace{0.1em}
{\footnotesize\texttt{sxw992@case.edu}\par}
\vspace{0.1em}
\vspace{0.8em}

\margintag{Two API versions}{Claude Code's agent-teams API changed with
v2.1.178, simplifying the team-registration layer. This manual covers both
versions, with the new API as the primary treatment; the core
argument---files over context, checkpoints, and one-command
reactivation---is unchanged by that simplification.}
\begin{tcolorbox}[
  enhanced,
  colback=burgunlight,
  colframe=burgundy,
  toprule=1pt, bottomrule=1pt, leftrule=0pt, rightrule=0pt,
  arc=0pt, boxsep=3pt,
  left=8pt, right=8pt, top=5pt, bottom=5pt,
  title={\small\bfseries\color{burgundy}Abstract},
  fonttitle=\small,
  attach boxed title to top left={yshift=-1mm,xshift=6pt},
  boxed title style={colback=white,colframe=white,sharp corners}
]
\small
Large Language Model (LLM) agents have significantly improved coding and
programming workflows, especially Claude Code, one of the most powerful LLM
coding agents, which conducts complex coding tasks. However, several drawbacks
can harm long-term agentic workflows: (1)~\textit{Irrecoverable agent
teams}---the Agent Teams feature is powerful, but the working state each
teammate builds up is lost and cannot be resumed once the process stops, e.g., a
closed terminal. (2)~\textit{Compaction erodes working detail}---compaction
condenses the conversation into a summary, so an agent's working details can
become fuzzy. (3)~\textit{Agentic ``technical debt''}---over time, a user's
decisions and the agents' operations get trapped in compacted old chats, and the
project becomes harder and harder to maintain and review. (4)~\textit{Heavy
prompt writing}---assigning or handing off tasks makes you rewrite long prompts
to achieve the expected agentic performance.
We propose \textbf{ATWZ} (\underline{A}gent \underline{T}eam \underline{W}ork
\underline{Z}one), a filesystem-based operations layer built around Claude
Code's native Agent Teams that addresses the pains above. Its central design
idea is to treat each agent and teammate as a human employee, and to keep their
important working state in files stored in a specific directory called a
`workstation', along with the skills, hooks, and scripts that utilize and
maintain these files. With ATWZ, an agent team can periodically back itself up,
so an agent's knowledge can be recovered after compaction; after a process ends,
the team can be restored with a single command; and these features also greatly
mitigate the ``technical debt'' above. Moreover, for communication within ATWZ,
agent ``employees'' can send documents to one another, which greatly reduces the
workload of writing prompts.
We present this as a developer-oriented design manual: we describe the design and
its rationale, illustrate it through scenarios and through using the layer to
build itself, and analyze its failure modes and limitations. We report no
benchmarks or user studies; empirical evaluation is left to future work. ATWZ is open-source and available at \url{https://github.com/SR-A-W/agent-team-work-zone}
\end{tcolorbox}

\margintag{Dogfooded}{ATWZ is developed by Claude Code agent teams that
themselves run on ATWZ---the layer is iterated on through its own use.}

\vspace{0.6em}
{\color{burgundy}\rule{\textwidth}{0.5pt}}
\vspace{0.4em}

\section{Introduction}
\begin{figure}[ht]
  \centering
  \includegraphics[height=0.6\textheight,keepaspectratio]{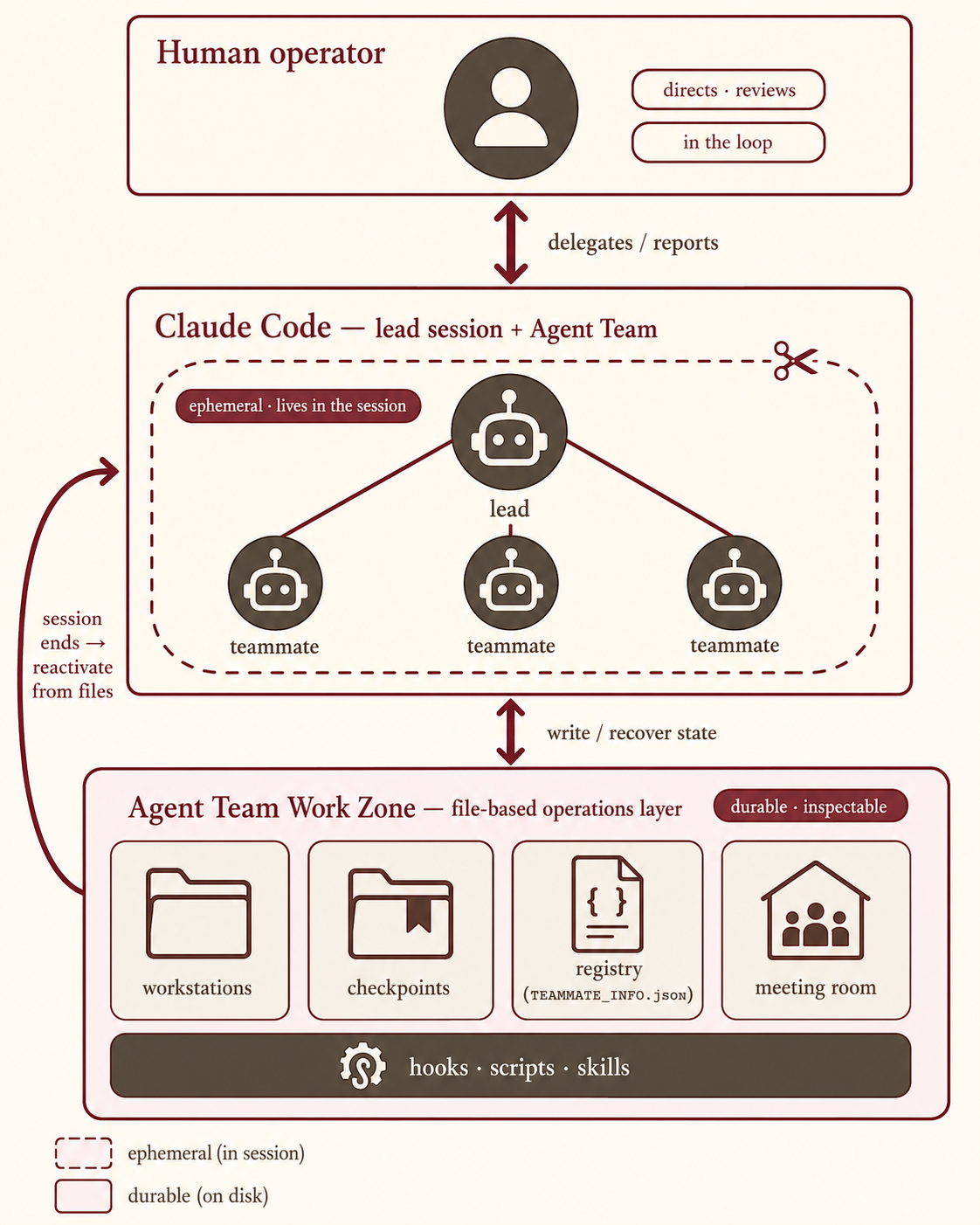}
  \caption{Agent Team Work Zone: a file-based operations layer over Claude Code.
    Agents remain ephemeral (middle); the durable workspace (bottom) lets an
    interrupted team be rebuilt.}
  \label{fig:overview}
\end{figure}

Coding agents are now capable enough to be run not one at a time but as a team:
a coordinator---a team lead---that decomposes a task and delegates to
specialists, which Claude Code supports directly. A team built this way inherits a property of the
underlying agent---its state lives inside a chat session. A teammate is a live
process with its own context window; when the session that owns it stops (an SSH
drop, a closed terminal, a restart), that teammate goes with it. The session can
be resumed (\texttt{/resume}), which restores the lead, but---as Claude Code's
documentation makes explicit---not its in-process teammates, which must be
rebuilt \citep{claudecode-agentteams}. Compaction---the summarization that keeps
a long conversation within the context window---erodes the same state more
gradually.

This is not specific to one tool: OpenAI Codex likewise offers only subagents
that it spawns for a task and closes on completion, with no team that persists
across sessions \citep{codex-subagents}. For short tasks this is fine. For
work that outlasts a single conversation it is less so: the human becomes the
team's memory and scheduler, decisions get buried in compacted chats, and it
grows unclear which agent did what---a kind of \emph{agentic technical debt}
that makes a project harder to resume and review.

This manual describes \textbf{Agent Team Work Zone}, a persistence and
management layer built around Claude Code's Agent Teams. Its one design choice
is to keep important state in files, outside the session's in-context
memory.\sidenote{\emph{In-context memory} = the session's live context window.
It is fast to read and write, but lost at session end or compaction---anything
not written to disk disappears with it.} Roles, working context, decisions,
messages, task files, and the team roster become durable, per-agent artifacts on
disk, tied to Claude Code by a thin layer of hooks, scripts, and skills. The
agents stay ephemeral---we do not change how Claude Code runs them---but what
they need in order to be reconstituted is written down. On this substrate the
layer adds automatic checkpoints, a one-command reactivation flow that rebuilds
a team from them, a file-based inbox for cross-session communication and audit,
and conventions for roles, versioning, and release, with the human in the loop
throughout. \Cref{fig:overview} shows the overall architecture.

Claude Code's agent-teams API changed with version~2.1.178, simplifying the
team-registration layer; this manual covers both versions, with the new API as
the primary treatment.  The core argument---files over context, checkpoints, and
a one-command reactivation flow---applies to both.

We present this as a developer-oriented design manual, not an empirical
study: we describe the design and its rationale, illustrate it with scenarios,
and report what we learned from using the layer to build itself. We make no
benchmarking or user-study claims; empirical evaluation is left to future work.
Sections~2--3 cover the background and the gaps we target; Sections~4--9
develop the design (version-specific differences are called out where they
arise); \Cref{sec:failures} catalogs failure modes and mitigations;
\Cref{sec:related} situates the work in related research; \Cref{sec:limits}
discusses limitations.

\section{Background: Claude Code, Subagents, and Agent Teams}%
\label{sec:background}

This manual builds directly on Claude Code's multi-agent features, so we first
describe how they work and, more importantly, where the state they produce is
lost. We then note that the same limitation appears in another widely used
coding agent, OpenAI Codex, which suggests the gap this manual addresses is not
a quirk of a single tool.

\subsection{Claude Code, subagents, and agent teams}

Claude Code is a command-line coding agent: a user converses with a model that
can read and edit files, run commands, and iterate toward a goal, all within a
single session. Two features let one session marshal more than one agent.

\emph{Subagents} are lightweight delegates. The main agent spawns one with its
own context window, tool access, and system prompt to handle a bounded piece of
work, and receives its result back; several can run in parallel. They are scoped
to a single session: the documentation states that ``[s]ubagents work within a
single session,'' and each runs in its own context window, returning its result
to the caller \citep{claudecode-subagents}.\sidenote{Session-scoped does not mean
nothing persists. The documentation records two opt-in forms of subagent
persistence: a subagent can be resumed with its full conversation history---within
a live session by messaging it, and after a restart only by resuming the same
parent session---and an optional \texttt{memory}
directory gives it a folder that ``survives across conversations.'' Neither makes
a subagent a live participant once its session is gone, which is the property
this manual is after; the \texttt{memory} directory is, in fact, the same move
this manual generalises---put the durable part in files. Documentation as of
July~2026.}

\emph{Agent Teams}, an experimental feature enabled by an environment flag, go
further. Several independent Claude Code instances are coordinated by one
\emph{team lead}; the \emph{teammates} ``work independently, each in its own
context window, and communicate directly with each other,'' while the lead
assigns tasks from a shared list and synthesizes results
\citep{claudecode-agentteams}. Claude Code offers other ways to run several
sessions at once---agent view dispatches independent background sessions---but
those workers ``report only to you''; Agent Teams are the only mode whose
workers are coordinated by a lead, share a session task list, and message one another
\citep{claudecode-agents}, and they are what this manual builds on.

\subsection{Where the state goes}

Both mechanisms make the session the boundary of durability for the agents
themselves---for subagents by default, with the opt-in escapes noted above, and
for agent teams as a documented limitation:
``\texttt{/resume} and \texttt{/rewind} do not restore in-process teammates.
After resuming a session, the lead may attempt to message teammates that no
longer exist. If this happens, tell the lead to spawn new teammates''
\citep{claudecode-agentteams}. The asymmetry is the point: a session can be
resumed (\texttt{/resume}), which restores the lead; its teammates, however, are
not restored and must be rebuilt. The teammate loss is version-independent:
they are live processes, and a session restart ends them regardless of release.

What happens to the team's \emph{on-disk} scaffolding---its config and task
list---depends on the version. On pre-2.1.178 releases, this on-disk state
persists past the session even though the runtime registration is gone: a dead
teammate's entry lingers in the team directory as a stale \emph{ghost} that
later runs must contend with (\Cref{sec:persistence}). On Claude Code
(hereafter CC)~$\geq$~2.1.178, this is replaced
by an automatic, session-scoped team (\texttt{session-<id>}) that is cleaned up
on exit: ``[t]he team config directory is removed when the session ends,'' while
``[t]he task list directory persists locally and is never uploaded, so resumed
sessions keep their tasks'' \citep{claudecode-agentteams}. (Documentation as of
July~2026; this wording changed during 2026.)\sidenote{On CC~$\geq$~2.1.178 the session-scoped
team also auto-cleans exited teammates, eliminating the ghost-entry and
\texttt{Name-2}-drift problems of the earlier API. Teammate \emph{processes},
however, remain ephemeral in both versions: a session restart still kills them
all. \Cref{sec:persistence} covers the two versions side by side.}

\begin{figure}[t]
  \centering
  \includegraphics[height=0.55\textheight,keepaspectratio]{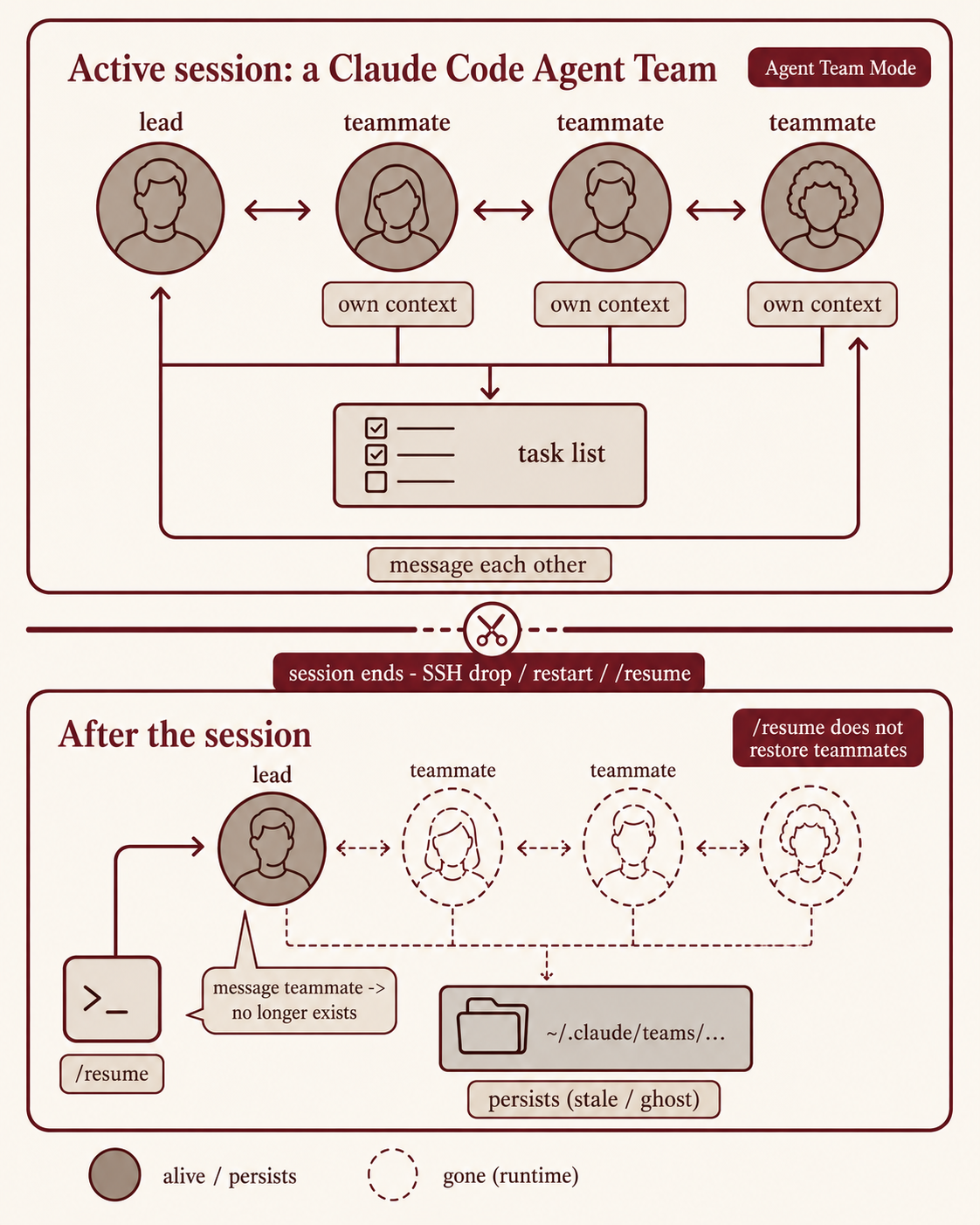}
  \caption{What a session end destroys (drawn for pre-2.1.178 releases): the
    teammate processes, the team's \emph{in-memory} coordination state, and any
    working context held only in a context window.  The lead's session can be
    \texttt{/resume}d; on pre-2.1.178 releases the on-disk team directory lingers
    as a stale ghost, whereas on CC $\geq$~2.1.178 the session-scoped team
    config is cleaned up on exit.  Either way the live team does not come back
    on its own.}
  \label{fig:ephemeral}
\end{figure}

So when a session stops, three things vanish together (\Cref{fig:ephemeral}): the teammate processes,
the team's in-memory coordination state, and any working context a teammate held
only in its context window. (A feature request to make subagent sessions
resumable was closed by an inactivity bot in January~2026 rather than by a
maintainer decision; session-scoped resumption has since shipped, but not the
cross-session interface the request asked for \citep{claudecode-issue7317}.) Compaction
is a softer form of
the same loss: it condenses a conversation into a summary to stay within the
context window, which can blur an agent's sense of its role, its current task,
and its outstanding commitments.\sidenote{\emph{Compaction} (\texttt{/compact})
rewrites the live context window as a condensed summary---fast but lossy. Role
definitions, nuanced decisions, and pending commitments are the most common
casualties.}

\subsection{The same shape in Codex}

This is not specific to Claude Code. OpenAI Codex exposes subagent workflows
in which the main agent spawns specialized agents in parallel and collects their
results in one response, and it closes the agent threads when they finish
\citep{codex-subagents}.\sidenote{Through mid-2026 the documentation stated that
``Codex only spawns subagents when you explicitly ask it to''; that documentation
has since moved to ChatGPT's site, where it now addresses ChatGPT and Codex
together, and as of July~2026 a higher tier ``enables proactive delegation, so
ChatGPT can delegate suitable independent work without a separate request.'' The change is in \emph{who
initiates} a subagent, not in its lifetime: the threads are still closed on
completion, and no standing team is introduced.} Its cloud mode runs
delegated \emph{tasks}, each in its own environment, that return a diff or pull
request \citep{codex-cloud}. In neither case is there a standing, named team
that survives across sessions; the only thing that persists is reusable agent
\emph{definitions} saved as on-disk configuration, not live agent instances.

In both tools, then, the multi-agent primitives are ephemeral: useful for
parallelizing a bounded task, but not for sustaining a team whose work spans
many sessions.

\subsection{The gap this creates}

The capability these tools provide---several specialized agents working under a
coordinator---is exactly what a non-trivial project wants. What they do not
provide is \emph{continuity}: a way for that team, and the understanding it
accumulates, to be carried past the end of a session. Closing that gap, without
changing how the agents themselves run, is the starting point for the rest of
this manual.

\section{Gaps and Problem Statement}

\Cref{sec:background} leaves a specific problem: Claude Code lets a team \emph{work}, but
not \emph{persist}. We target the gaps below---each a place where unaided
Claude Code loses continuity that a longer-lived project needs.

\begin{itemize}
  \item \textbf{Teammates do not survive a restart.} When the lead session ends,
    teammates and their working state are gone, with no native way to bring them
    back; the user rebuilds the team and re-explains the context.
  \item \textbf{Compaction loses detail.} A summarized conversation can blur an
    agent's role, current task, and outstanding commitments.
  \item \textbf{Agents in different sessions cannot talk.} Separate conversations
    have no channel to one another, and in-team messages are not retained after
    the session.
  \item \textbf{Task and decision state does not live with the project.} The
    platform's task list is not deleted when the session ends, but it sits
    outside the project tree and records task items only: what was decided and
    why, and what was finished, leave no durable, inspectable trace beside the
    work itself, so reasoning is duplicated or dropped and changes are hard to
    attribute. Even that task list is swept on a retention timer---the same
    \texttt{cleanupPeriodDays} that clears session transcripts, thirty days by
    default~\citep{claudecode-settings}.
  \item \textbf{Hand-offs are re-typed.} Passing a task across a conversation
    boundary means rewriting its background, goal, and constraints by hand.
\end{itemize}

The platform's own persistence makes the shape of the gap unusually clear. On
CC~$\geq$~2.1.178 a resumed session recovers its session task list---the work items
survive---while its in-process teammates, as \Cref{sec:background} noted, do
not \citep{claudecode-agentteams}. What returns is the \emph{run}: a list of
things to do, with no one left to do them. What does not return is the
\emph{team}---its roles, its accumulated understanding, its outstanding
commitments---and making that durable is what the rest of this manual is about.

We address these at the level of the \emph{workspace}, not the model. We do not
modify Claude Code's agents or propose a new agent algorithm; the contribution
is a set of file-based conventions, with supporting hooks, scripts, and skills,
that give a Claude Code agent team continuity, recoverability, and an audit
trail. One boundary is worth stating plainly: we do not keep teammate processes
alive across a session, and we claim no platform-level persistence the tool does
not provide. Recovery here is \emph{reconstruction} from durable
files\sidenote{Reconstruction, not preservation: a fresh teammate re-reads what
the previous one wrote down and resumes from that state. No live process is
carried over.}---a fresh teammate re-reads what the previous one wrote
down---not preservation of a live process.

This invites a fair objection: if a lead resumes its own session and then rebuilds
its teammates, is that not what resuming a subagent inside its parent session
already does? The ritual is indeed similar; three things differ. The unit is a
team rather than a single agent---a roster with roles, scopes, and outstanding
commitments, for which the platform offers no equivalent. The artifact is authored
rather than captured: a checkpoint is a summary its writer composed for its
successor, deliberately lossy and legible to a person, not a raw transcript. And
the artifact lives in the project, so it is versioned, reviewable, and independent
of any platform retention window---resuming the original session is a convenience
for the lead, not a precondition, since reactivation reads the registry and the
workstations, both on disk, so a fresh session in the same project can rebuild the
team.

\section{Design Philosophy}

\subsection{An operations layer, not a replacement}

\begin{figure}[t]
  \centering
  \includegraphics[height=0.58\textheight,keepaspectratio]{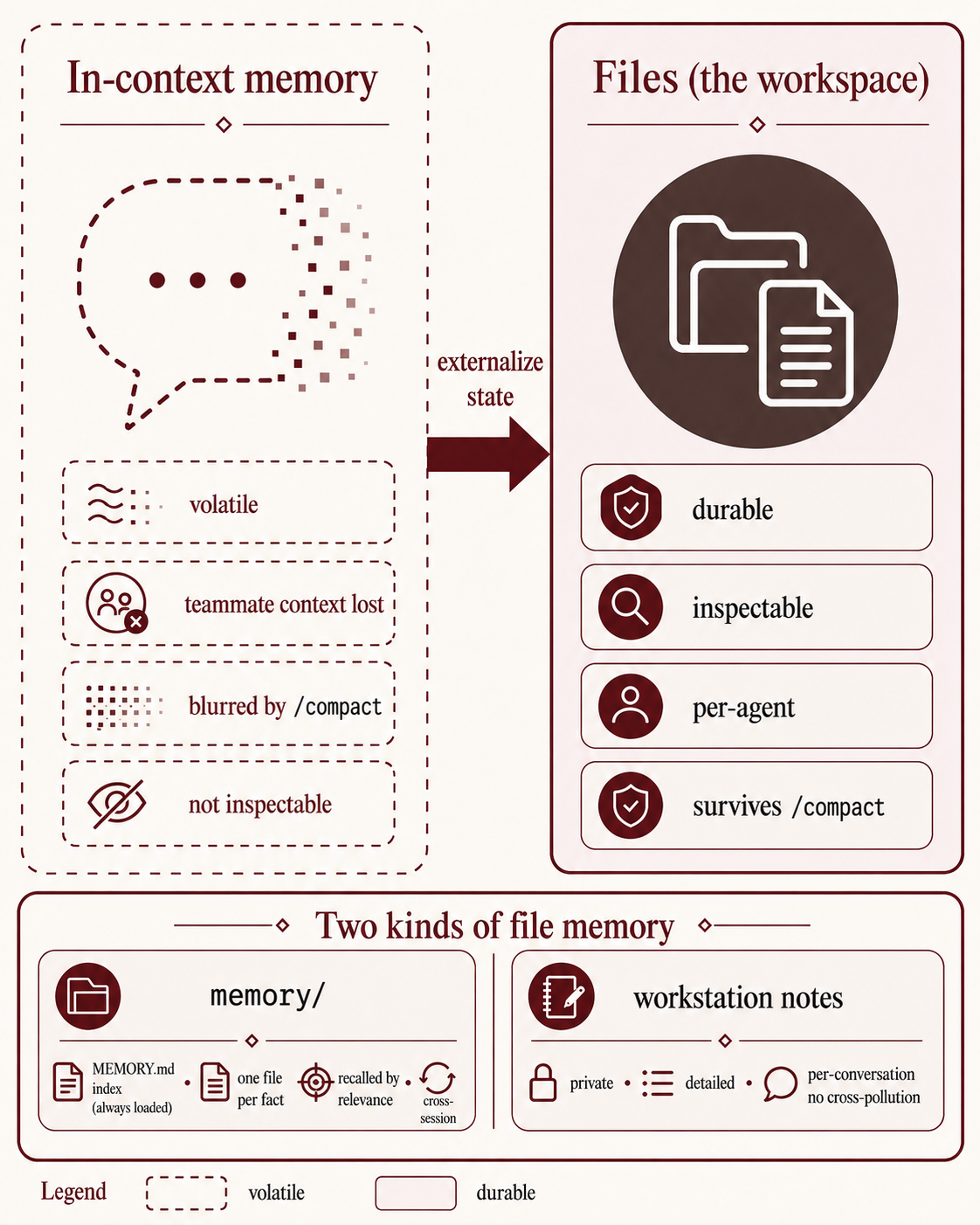}
  \caption{Keeping state in files rather than in the volatile, in-context
    memory; and how a per-agent workstation note differs from Claude Code's
    relevance-recalled \texttt{memory/}.}
  \label{fig:files-vs-context}
\end{figure}

Agent Team Work Zone does not change how Claude Code's agents run; it wraps
them in a file-based operations layer. It assumes the platform's
strengths---file I/O, project-local skills and custom agents, hooks, and Agent
Teams---and adds what a longer-lived project needs around them. Put bluntly:
Claude Code lets you \emph{generate} agents; this layer helps you \emph{manage}
them after the conversation ends. Everything follows from one rule---anything a
future agent will need is written to a file, outside the session's in-context
memory (\Cref{fig:files-vs-context}).

\subsection{State in files, not in-context memory}

By \emph{in-context memory} we mean the live context window: what the model is
currently holding in the conversation. It is fast and convenient but
volatile---it ends with the session and is thinned by compaction---and it cannot
be inspected from outside. For teammates this loss is permanent (they cannot be
resumed); a lead may recover its own context with \texttt{/resume}. The design
treats it as working space, not storage.
Anything that must outlast a teammate---its role, what it is doing, what it
decided, what it still owes---is written to disk instead.

This is distinct from Claude Code's always-loaded instructions, \texttt{CLAUDE.md}.
Those live in up to four scopes---managed, user, project, and local---which are
``concatenated into context rather than overriding each other,'' and can be
extended with path-scoped rules under \texttt{.claude/rules/}
\citep{claudecode-memory}: handy for stable, global instructions, but a shared
surface, automatically consumed by every session. Nor is it Claude Code's recall-based memory: that system surfaces
facts by relevance through a \texttt{MEMORY.md} index over one-file-per-topic
notes, rather than loading everything up front~\citep{claudecode-memory}. A
workstation note is different from both: private to one agent, detailed and
specific to its current task, read on demand rather than forced into every
context window, and isolated so that one agent's notes never overwrite or
pollute another's. The three are complementary---always-loaded instructions in
\texttt{CLAUDE.md}, relevance-recalled facts in Claude Code's \texttt{memory/},
and per-agent operational state in workstations.

\subsection{Principles}

Six principles follow from the file-first rule and recur throughout the design.

\begin{itemize}
  \item \textbf{Files over in-context memory.} If a future agent will need it,
    write it down. This is the root rule; every other primitive is a consequence
    of it.
  \item \textbf{Workstations as working notes.} Each agent keeps an external
    notebook---current understanding, dead ends, open commitments, where to
    resume---whose real job is reconstruction after compaction or a restart.
  \item \textbf{Low coupling.} Each agent owns its workstation; others interact
    only through reports, never by editing another agent's files.
  \item \textbf{Reports as prompts.} A good agent-to-agent report \emph{is} a
    prompt packet: what happened, what was tried, what changed, what is needed
    next, and where the files are.
  \item \textbf{Leads coordinate.} A team lead spends its context on
    decomposition, routing, review, and synthesis; implementation goes to
    teammates.
  \item \textbf{Auditability.} Finished work and the reasoning behind it stay on
    disk, so a project's history does not vanish into closed conversations.
\end{itemize}

\subsection{Two forms, not one}

Not every agent becomes a team. Each teammate runs in its own session, so token
cost grows with team size. Simple or coordination-only work stays \emph{flat}---
one agent, one workstation; only genuinely complex work, with several specialties
or parallel streams, is given a \emph{team}. The structure mirrors the
restraint: flat and team workstations sit side by side---a team workstation being
one whose directory carries the \texttt{\_team} suffix \emph{and} the
team-internal subtree (\texttt{roundtable/}, \texttt{teammates/}, the registry)---
so the whole roster is visible at a glance.  Promoting a flat agent is that
rename plus the creation of that subtree, which \texttt{/promote-to-team}
performs in one step.

\subsection{The layer is hardened by use}

A note on method rather than mechanism. Much of this design did not come from
up-front planning; it came from running the layer on real work---including
building the layer itself---and hardening the failures that surfaced into rules
and mechanisms. Many of the sharpest problems we hit only appeared under
sustained use, so we treat ``use it on itself, then fix what breaks'' as part of
the method. \Cref{sec:failures} catalogs what that produced.

\section{Workspace and Core Primitives}%
\label{sec:primitives}

This section describes the workspace the layer installs and the handful of
primitives the rest of the design is built from.

\subsection{Workstations}

The basic unit is a \textbf{workstation}: a persistent directory owned by
exactly one agent, holding everything that agent knows, owes, or is
tracking---as files, not chat memory. There are two kinds, distinguished by name
and structure together. A \emph{flat} workstation is a plain directory; a
\emph{team} workstation ends in \texttt{\_team} \emph{and} carries the
team-internal subtree, both of which promotion creates in one step. Skills tell
them apart at
runtime by structure alone---the directory ends in \texttt{\_team} and contains
a \texttt{roundtable/}---so no configuration file or flag is needed; the layout
itself is the signal.

\subsection{The directory tree}

Everything installs under \texttt{\_agent\_team\_work\_zone/} in the project
root, keeping framework-owned files separate from user-owned ones:

\begin{center}
\begin{minipage}{0.85\linewidth}
{\small\ttfamily
\begin{verbatim}
_agent_team_work_zone/
|-- README.md            framework doc + work rules (in marked blocks)
|-- meeting_room/        top-level async comms (all flat agents + leads)
|-- archive/             resolved meeting-room messages
|-- <agent>/             FLAT workstation
|     \-- README.md  notes.md  TODO.md  ACTIVE_JOBS.md  COMPLETED_JOBS.md
|           (README.md: role + the full work rules)
|-- <agent>_team/        TEAM workstation
|     |-- README.md  ...  TEAMMATE_INFO.json   (registry; role + full rules)
|     |-- roundtable/    team-internal comms (lead <-> teammates)
|     |-- archive/       team_recipes/
|     \-- teammates/<name>/
|           \-- README.md  working-context.md  completed.md
|                 TODO.md  commitments.md
|                 (README.md: role + the condensed rule subset)
|-- resources/           framework-owned: skills, agents,
|                          role archetypes, scripts, hooks
\-- docs/                framework-owned: the manuals
\end{verbatim}
}
\end{minipage}
\end{center}

The split is load-bearing for upgrades. As of release~v0.3.2, an upgrade
replaces the framework-owned paths---\texttt{resources/}, \texttt{docs/}, and the
comment-marked blocks of \texttt{README.md} (its framework section, its work
rules, and its reference material: the pre-installed skill, subagent, and
troubleshooting listings)---while user-owned paths (every workstation, the
meeting room, and the archive) are never touched. The boundaries inside
\texttt{README.md} are enforced mechanically by comment markers, so an upgrade
rewrites only those blocks and leaves project-specific content verbatim
(\Cref{sec:versioning}).

\subsection{Roles, the registry, and communication}

Three primitives that recur in later sections are introduced here. A
\textbf{role} is a short definition of what an agent is for; it is stored as a
file, so a future spawn of the same agent inherits it rather than being
re-explained. The \textbf{team registry}---\texttt{TEAMMATE\_INFO.json}, owned by
the lead, who adds and removes entries, while each teammate stamps its own
checkpoint time into its row---lists who is on the team and the metadata that
drives recovery; it is the input to reactivation (\Cref{sec:persistence}). And agents coordinate
through two tiers of file-based, asynchronous \textbf{communication}: a
top-level \emph{meeting room} shared by all flat agents and leads, and a
per-team \emph{roundtable} for a lead and its teammates, detailed in
\Cref{sec:comms}. Each of these is just files in the tree above---which is the point.

\section{Persistence and Recovery}%
\label{sec:persistence}

Carrying a Claude Code agent team past the session that created it is the
layer's signature problem.  Two facts make it hard---Claude Code does not
persist teammate sessions, and judging whether a teammate is still alive is
easy to get wrong.  The design answers the first with a reactivation procedure
built on a durable registry and on checkpoints, and the second with an explicit
liveness rule.  The mechanics of session-state management changed with
Claude Code~2.1.178; this section covers both versions, new API first.

\subsection{Team state across a session boundary}

\paragraph{New version (CC $\geq$~2.1.178).}
Claude Code creates one automatic, session-scoped team---named
\texttt{session-\allowbreak<id>}---per lead session.  When a teammate
leaves, the runtime keeps the registration in sync: no ghost entry accumulates
and the team's membership record stays current.  After a session
boundary the registration is recreated automatically in the next lead session,
so there is only one thing the operator needs to rebuild: the teammates
themselves.  The platform's own task list is unaffected by the boundary: it is
stored under a session-derived name and its directory is not removed when the
session ends, so a resumed session recovers its tasks.

\paragraph{Prior versions (pre-2.1.178).}
Earlier releases kept team state in two places that behave oppositely across a
session boundary.  \emph{Runtime registration} lives in process memory and is
destroyed when the session ends; afterwards, any attempt to spawn into the team
fails (``Team does not exist'').  The on-disk \emph{team config} persists---and
accumulates: a dead teammate leaves a ``ghost'' entry that is never cleaned up,
so re-spawning the same name yields a drifted \texttt{Name-2}.  The two fail in
opposite directions---one is gone and must be rebuilt, the other lingers and
must be cleared---so recovery clears and rebuilds both at once.

\subsection{Reactivation}

\paragraph{New version (CC $\geq$~2.1.178).}

\begin{figure}[t]
  \centering
  \includegraphics[height=0.58\textheight,keepaspectratio]{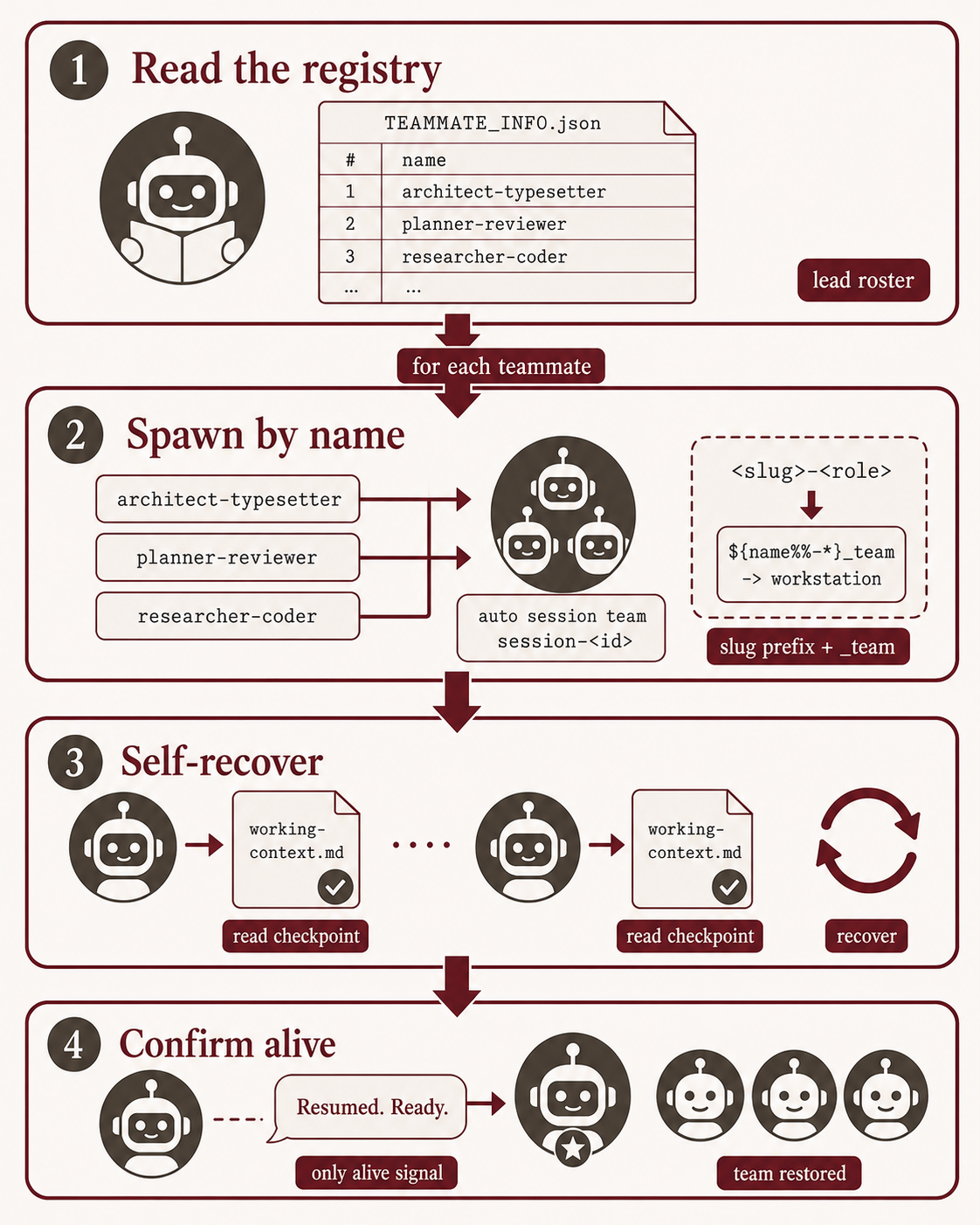}
  \caption{Reactivating a team on CC $\geq$~2.1.178: read the team registry,
    spawn each teammate by \texttt{<slug>-<role>} name into the automatic
    session team, and let each teammate recover from its own checkpoint.
    No registration-clearing step is needed.}
  \label{fig:reactivation-new}
\end{figure}

\texttt{/reactivate-team} on the new API has no clearing step
(\Cref{fig:reactivation-new}): the session team
is already fresh, so recovery reads the team registry
(\texttt{TEAMMATE\_INFO.json}, the lead-owned roster) directly, and for
each teammate spawns a fresh session whose opening instruction is to read its
own workstation---its role file, its checkpoint, its commitments.  The layer's
skills name every teammate in the \texttt{<slug>-<role>} convention (see
\Cref{sec:roles}): the idle hook derives the workstation path from the slug
prefix, which is what makes that resolution unambiguous---the name carries
structure, not just identity.  Success is judged one
way only---the revived teammate sends a message back in the current session.
Disk files cannot confirm a particular reactivation, since a spawn that dies
before messaging leaves every file unchanged; the receipt, not the artifact, is
the evidence.

\paragraph{Prior versions (pre-2.1.178).}

\begin{figure}[t]
  \centering
  \includegraphics[height=0.58\textheight,keepaspectratio]{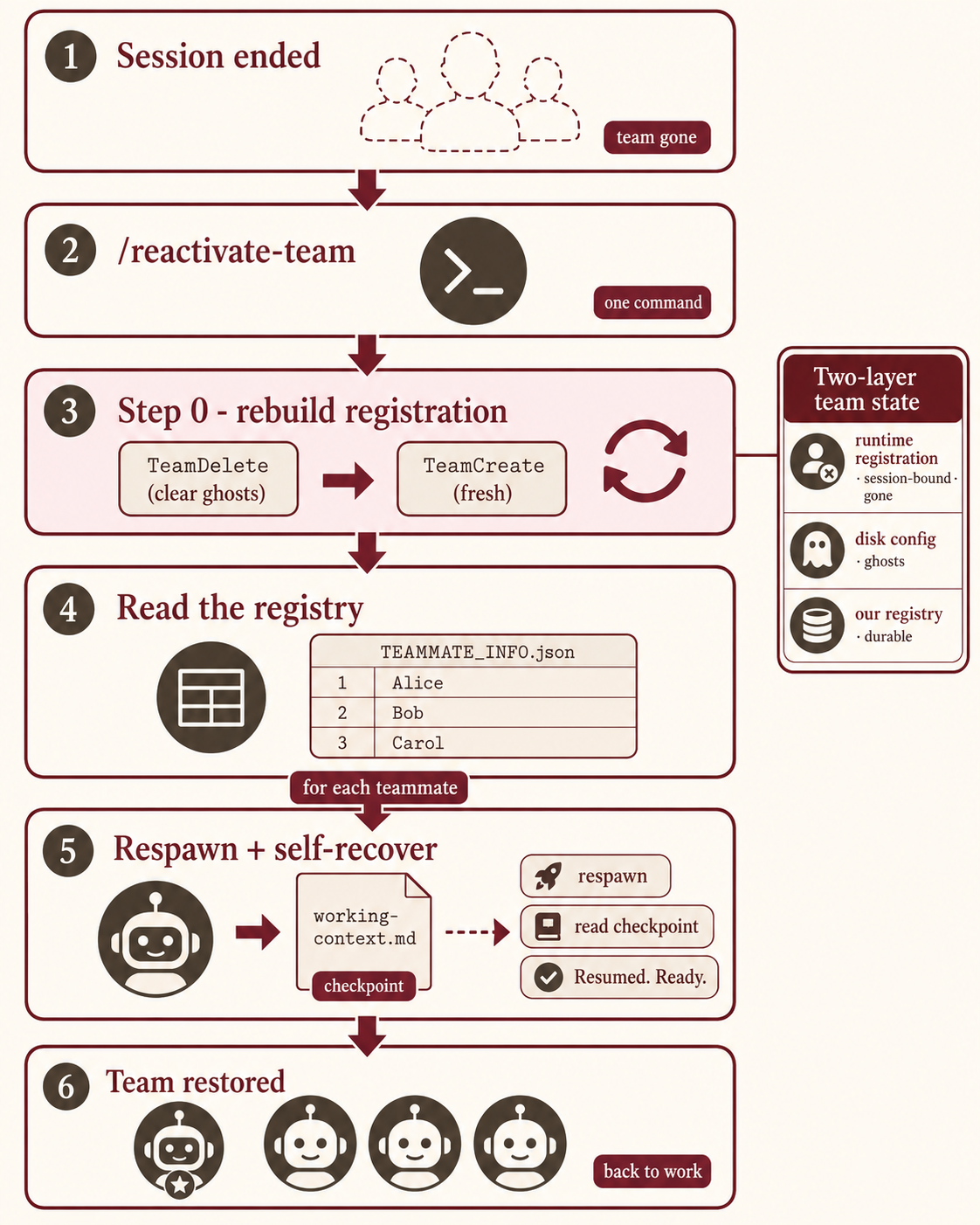}
  \caption{Reactivating a team on prior versions: clear the stale registration,
    read the registry, and respawn each teammate so it recovers from its own
    checkpoint.}
  \label{fig:reactivation}
\end{figure}

On earlier releases (\Cref{fig:reactivation}) \texttt{/reactivate-team} must
first clear the stale
config and recreate the registration in a single step---so respawns reclaim
their original names rather than drifting to \texttt{Name-2}---and only
then reads the team registry and spawns each teammate with the same
read-your-own-notes opening instruction.  The rest of the flow is identical:
each teammate reads its own workstation and messages the lead to confirm it is
alive.

\subsection{Knowing whether a teammate is alive}

A recurring way to get recovery wrong is to \emph{assume} a teammate is alive
when it is not.  The framework fixes a single rule: a teammate is alive only if
it was spawned by the current, uninterrupted lead session; any restart kills
every teammate of that team---a resume restores the lead but not its teammates
(\Cref{sec:background}).
Static signals are explicitly not
evidence---not a config entry, not a message sitting in an inbox, not a
\texttt{status:~active} field a human wrote, and not a plain reply (which never
crosses the agent boundary).  The one positive signal is a message received
from the teammate this session, in reply to a fresh ping; its absence means
``unknown,'' which is treated as dead.\sidenote{On CC
$\geq$~2.1.178, teammate messages arrive tagged with the automatic session-team
name (\texttt{@session-\allowbreak<id>}); this is normal routing, not a
signal that the teammate is dead or has joined a different team.}

Two corollaries, both learned in use: context compaction happens in the same
process and does \emph{not} kill teammates, so it must not trigger a blind
rebuild; and a receipt expires---a teammate that answered several turns ago may
be gone now, so every judgment needs a fresh check.  A lighter third state,
\emph{benched}, lets a teammate be taken offline to free a slot with its
workstation preserved, then woken by name later.

\subsection{Checkpoints: automatic and non-destructive}

Reactivation is only as good as the state a teammate left behind, and that
state should not depend on anyone remembering to save it.  Each teammate keeps
a checkpoint file: a current-state snapshot (objective, active task,
decisions, open questions, commitments, key file references) together with a
short, append-only journal of recent work whose newest entry preserves the
last few turns verbatim while older entries are condensed to
summaries---keeping the log compact without losing recent precision.  This
file is read back on reactivation to reconstruct the teammate.  Writing a checkpoint is non-destructive: unlike
compaction, which rewrites the live context, a checkpoint only \emph{reads}
the context and \emph{writes} a file, so a teammate can checkpoint mid-task and
keep working.

\begin{figure}[t]
  \centering
  \includegraphics[height=0.58\textheight,keepaspectratio]{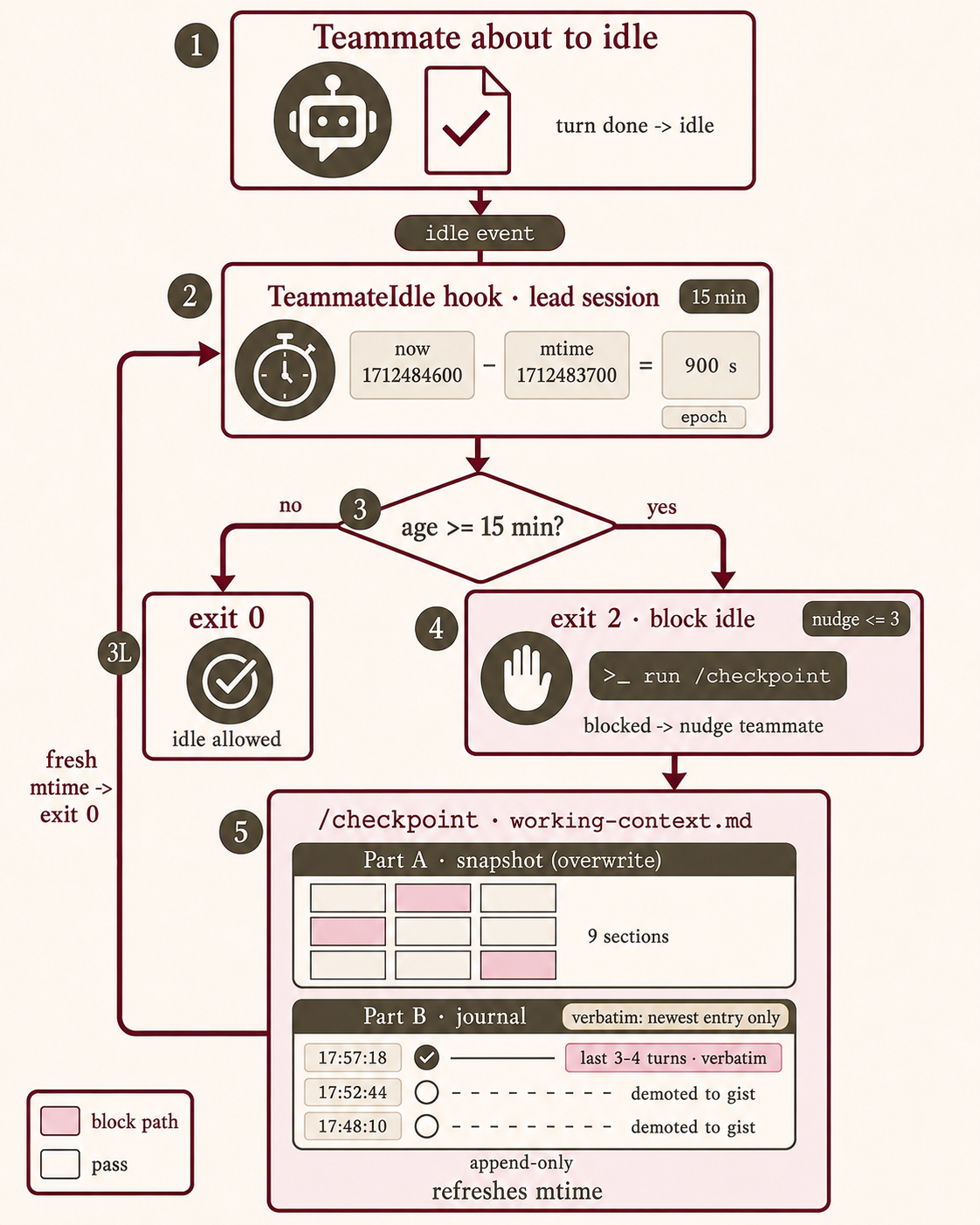}
  \caption{The \texttt{TeammateIdle} checkpoint gate: an age-based hook blocks
    a teammate's idle and forces a checkpoint, then self-brakes once the
    timestamp refreshes.}
  \label{fig:checkpoint}
\end{figure}

Saving is made automatic by a one-sided hook (\Cref{fig:checkpoint}).
When a teammate is about to go idle, a \texttt{TeammateIdle} hook in the lead
session checks how long ago that teammate last checkpointed, comparing Unix
epoch timestamps---plain integers, which sidesteps a class of timezone bugs.
Below a threshold (15~minutes) idle proceeds; at or above it, the hook returns
a non-zero exit that blocks the idle and feeds the teammate a reminder to
checkpoint now.  Writing the checkpoint refreshes the file's timestamp, so the
next idle passes and the loop brakes itself.  A guard releases a teammate after
a few consecutive nudges rather than wedging it, and any
uncertainty---a missing file, a tooling gap---fails open, never blocking a
teammate out of doubt.  The release gate (\Cref{sec:versioning}) defaults the
other way, and deliberately so: the same rule---fail toward the recoverable
outcome---points in opposite directions when the costs are reversed.  A missed
checkpoint costs at most a few minutes of unwritten context and the next idle
nudges again, whereas a wrongly blocked teammate stops working; a delayed tag
costs a rerun, whereas a bad release cannot be recalled.

\section{Communication and Audit}%
\label{sec:comms}

Agents in different sessions cannot talk to each other directly, so the layer
gives them an asynchronous, file-based channel---and a discipline for using it
that also leaves an audit trail.

\subsection{Two channels, one protocol}

Communication runs over the two file-based spaces from \Cref{sec:primitives}: a top-level
\emph{meeting room} that any flat agent or team lead can use, and a per-team
\emph{roundtable} for a lead and its teammates. Both run the same protocol,
differing only in how agents are addressed. A message is a markdown file whose
frontmatter carries its lifecycle---a status that moves \texttt{OPEN} ->
\texttt{IN\_PROGRESS} -> \texttt{RESOLVED}, a \texttt{from} (the issuer), a
\texttt{to} (one name, several, or everyone), and an optional read-only
\texttt{cc}. In the meeting room agents are named directly; inside a team they
are addressed as \texttt{<team>/<role>}, and a \texttt{kind} field records the
report type. The filename encodes author, type, and a minute-precise timestamp,
so messages sort chronologically and their authorship is visible without opening
them.

\begin{figure}[t]
  \centering
  \includegraphics[height=0.58\textheight,keepaspectratio]{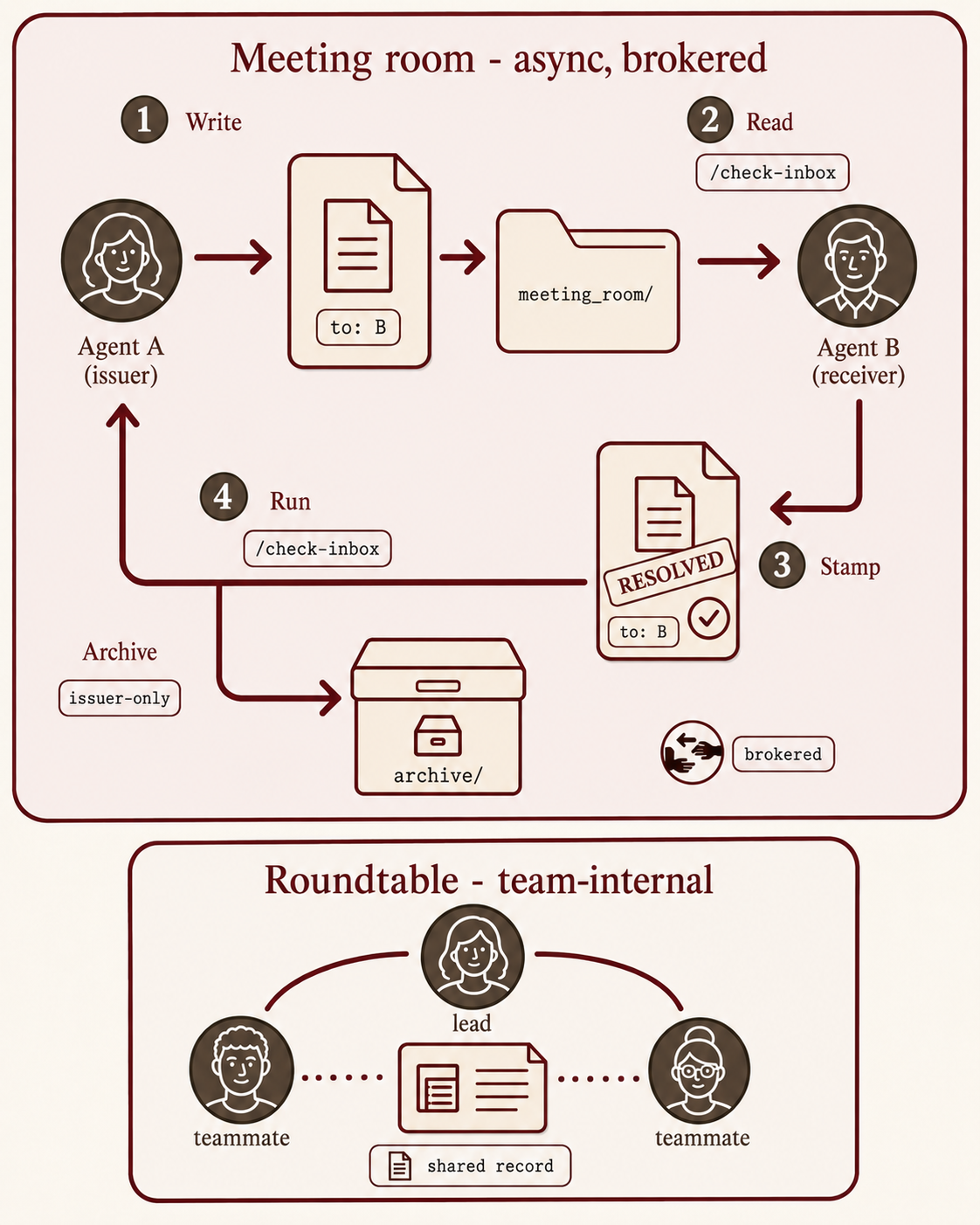}
  \caption{File-based messaging: an agent writes a message addressed to
    another; the recipient reads it with \texttt{/check-inbox}, does the task
    and marks it resolved; the issuer later archives it.}
  \label{fig:comm}
\end{figure}

\subsection{Processing an inbox}

Because the channel does not push, an agent reads it on demand with
\texttt{/check-inbox} (\Cref{fig:comm}). The skill works out the agent's
identity, scans the spaces that identity may read, sorts the actionable messages
oldest-first---a later task may depend on an earlier one---and presents them for
the user to confirm before anything runs. It then works each item (mark in
progress, do the task, record the result, mark resolved) and finally archives
the agent's \emph{own} completed messages.

\subsection{Why the issuer archives}

One small rule prevents a common failure: only a message's \emph{issuer} may
archive it. An earlier design let recipients archive, which caused two
problems---with several recipients, each assumed another would do it, so
resolved messages piled up; and a recipient could archive before the issuer had
seen the completion, losing the signal. Making archival the issuer's exclusive
duty leaves exactly one accountable party, who by definition has seen the result
first; messages with several recipients are coordinated by a checklist the
issuer clears. The archive itself is never pruned: it is the durable record of
what was decided and done, which is what keeps the work auditable after the
conversations that produced it have closed.

\subsection{Handing work across a boundary}

When a task must move from one agent to another, \texttt{/handoff} packages it:
progress, files touched, decisions, open questions, and---the thing most easily
lost in a transfer---\emph{why the task exists}. Its suggested next steps are
explicitly non-binding; the receiver chooses its own method. The originator's
task file marks the item as handed off rather than deleting it, so the trail
remains intact.

\section{Roles, Lifecycle, and Work Rules}%
\label{sec:roles}

A team is a lead plus a few specialists---formed and maintained through a small
set of skills, drawing on a library of roles, and governed by a short list of
work rules. This section covers that management machinery and the discipline
behind it.

\begin{figure}[ht]
  \centering
  \includegraphics[height=0.58\textheight,keepaspectratio]{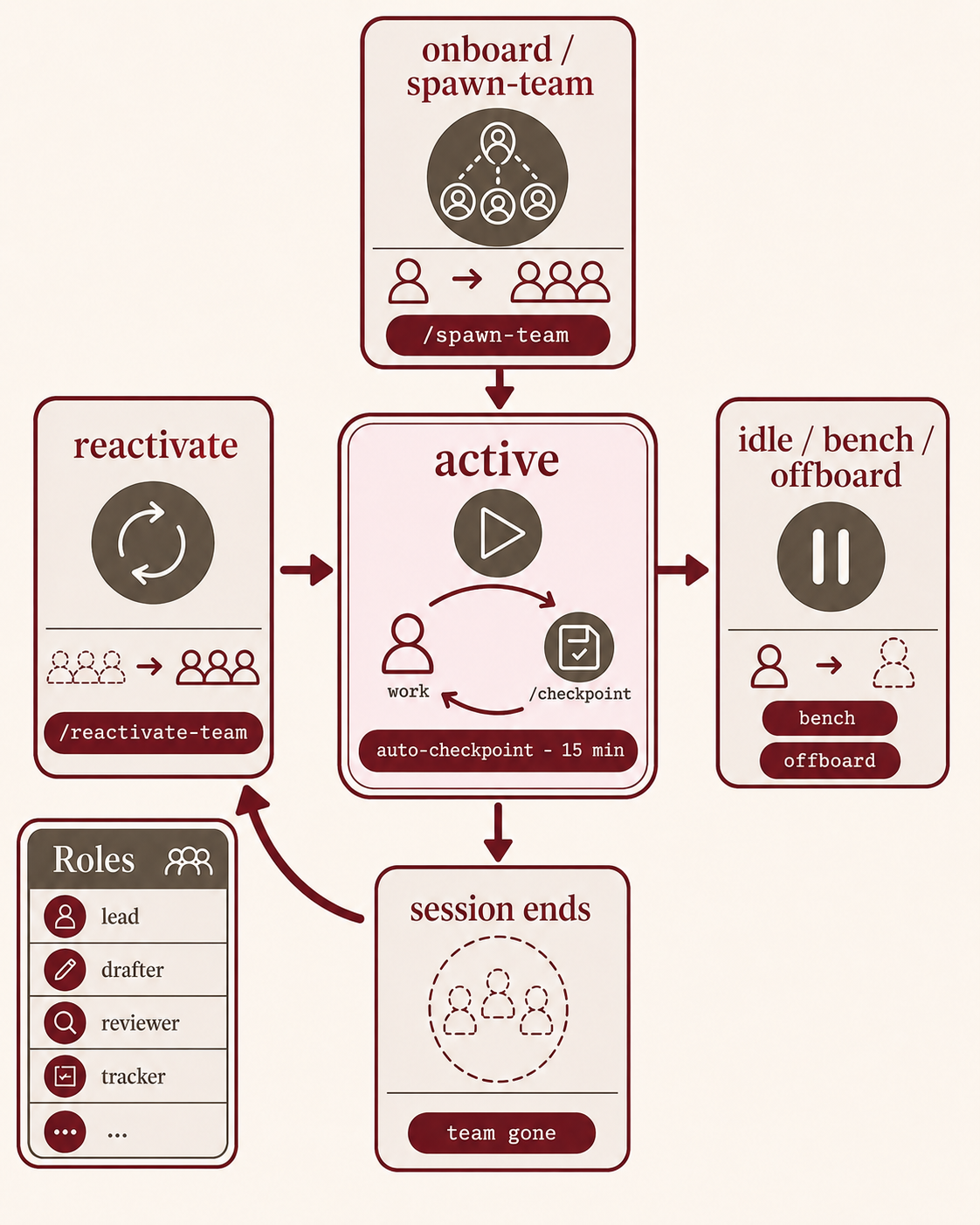}
  \caption{A team's lifecycle (baseline, shared across versions): spawn, an active
    loop of work and automatic checkpoints, bench or offboard, and reactivation
    after a session ends; plus the role roster of a lead and specialists (role names in the figure are illustrative).  See
    \Cref{fig:lifecycle-new} for the one addition specific to CC $\geq$~2.1.178.}
  \label{fig:lifecycle}
\end{figure}

\subsection{A small library of roles}

Roles come at two layers. Five generic \emph{subagents} ship with the layer and
are reusable as-is: a \textbf{tracker} for cheap periodic status polling, a
\textbf{reviewer} for read-only graded code review, an \textbf{investigator}
for hypothesis-driven debugging of ``ran fine but the result is wrong,'' a
\textbf{devil's-advocate} for adversarial review of a plan before it runs, and
a \textbf{git manager} for version-control operations with safety checks. The
model assigned to each reflects its work---mechanical polling on the cheapest
capable model, structured review a step up, and adversarial or
hypothesis-generating work (where a confidently wrong cheap answer is worse than
honest uncertainty) on the strongest. Beyond these, nine \emph{archetypes} are
templates rather than ready agents: ``coder'' is too vague to define globally,
so an archetype captures a stable category and a starting prompt---including an
explicit ``what you do \emph{not} do,'' to keep parallel teammates from
overlapping---which the lead fills in per task.

\subsection{Lifecycle}

A workstation begins with \texttt{/onboard}, which asks, rather than guesses,
whether the agent is flat or a team lead; a flat workstation can later be
promoted in place when its task outgrows it. A team is formed by
\texttt{/spawn-team}, a short flow that collects the task, decomposes it,
proposes a lineup for the user to confirm, then spawns the teammates and records
a \emph{recipe} of what it did. Once running, a lead grows, reviews, retires,
or temporarily \emph{benches} teammates through a handful of management skills
(\Cref{fig:lifecycle}).
Team size is kept small---three to five---since coordination cost grows faster
than linearly; offboarding is a neutral, normal step, and a retired teammate's
workstation is preserved as a record. After a session ends, teammates return to
the active state through \texttt{/reactivate-team} (\Cref{sec:persistence}).
Two recurring practices are worth naming:
gating a teammate's larger changes behind a quick read-only plan the lead
approves, which turns an expensive wrong edit into a cheap review; and an
optional adversarial reviewer during team formation, chartered to surface failure
modes but never to veto---the decision stays the lead's.

From the user's seat, most of this is invisible: the management skills are
agent-invocable, so the user gives a task in natural language and the lead
decides whether to form a team, whom to add, and whom to bench.

\paragraph{Naming and permissions (version callout).}

\begin{figure}[ht]
  \centering
  \includegraphics[height=0.58\textheight,keepaspectratio]{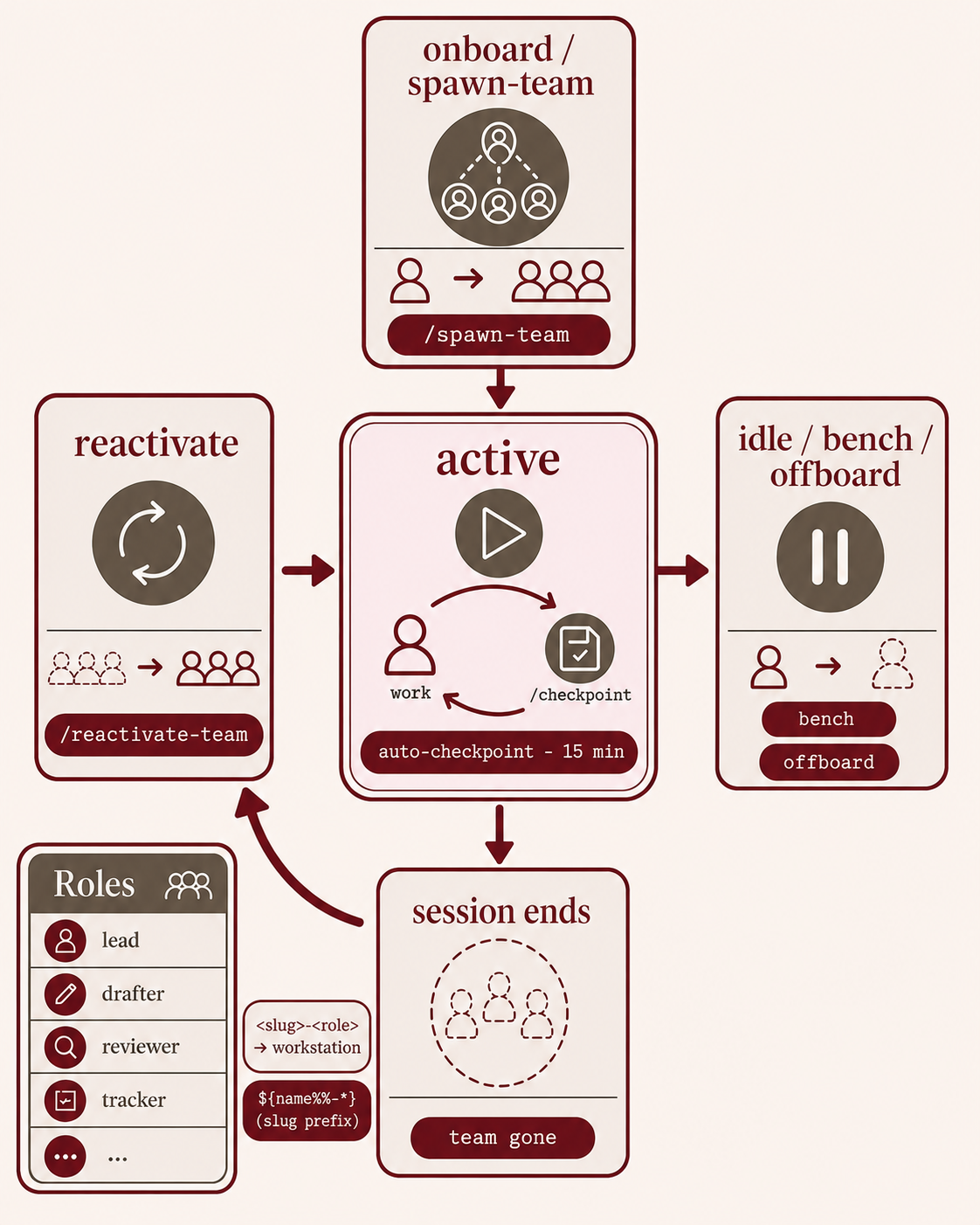}
  \caption{Lifecycle on CC $\geq$~2.1.178: \texttt{<slug>-<role>} is the naming
    convention the skills apply, letting the idle hook derive each teammate's
    workstation path from the slug prefix without ambiguity; on prior versions
    the hook read it from the idle event's team name instead.}
  \label{fig:lifecycle-new}
\end{figure}

On CC $\geq$~2.1.178, the layer's skills name every teammate in the
\texttt{<slug>-<role>} convention (for example, \texttt{architect-typesetter}).
The convention is not cosmetic: it lets the idle hook resolve a teammate's
workstation deterministically, deriving the path from the slug prefix at runtime
(\texttt{\$\{name\%\%-*\}\_team}, \Cref{fig:lifecycle-new}).  A legacy name
still resolves, through a
guarded glob over the workstation tree---but that fallback abstains when the name
is not unique across teams, silently dropping the nudge for both
(\Cref{sec:failures}).  On prior versions the hook resolved the workstation from
the idle event's team name directly, so any teammate name worked.

Permission mode is inherited from the lead session in both versions, rather than
set per teammate at spawn; to run teammates in \texttt{auto} mode, set
\texttt{"permissions": \{"defaultMode": "auto"\}} in the \emph{global}
\texttt{\textasciitilde/.claude/settings.json} before spawning---Claude Code
ignores this key at the project and local levels.

\subsection{Work rules}

Underneath the skills is a short list of behavioral rules, and delivering them is
treated as a mechanism rather than an exhortation. A flat workstation's or a
lead's role file carries the full set verbatim, and an upgrade refreshes it in
place, so that re-reading that file after compaction reloads the rules as they
now stand. A teammate's role file carries a condensed subset---the ones that
govern its own conduct---which \texttt{/spawn-team} writes in when it creates
the workstation, with a self-heal step in the spawn and reactivation prompts
for workstations created before this version. The rules encode disciplines this manual has already touched: each
agent owns its own files and never edits another's, acting only through
messages; reports must stand on their own; only an issuer archives; task state
and working notes live in files beside the project rather than the session's
task list; a lead spends its context on coordination, not implementation; and a
teammate keeps its checkpoint current. The list is less a
rulebook than the written form of the file-first philosophy applied to everyday
conduct.

\section{Versioning and Release}%
\label{sec:versioning}

The layer is a template that downstream projects install and later upgrade, so
its deployment has one hard requirement: an upgrade must refresh the framework
without ever touching the user's own work.

The version-specific behavior described in \Cref{sec:persistence,sec:roles} maps
onto a single linear release line, cut at the Claude Code~2.1.178 API change.
Release~v0.1.0 is the sole---and final---release for the prior API
(CC~$\leq$~2.1.177): that edition is frozen and takes no further changes.
Everything from~v0.2.0 onward targets the new API (CC~$\geq$~2.1.178) and is where
development continues; v0.3.2 is current as of July~2026. A project on an older
runtime installs the frozen release and stays there, while a project on the new
runtime upgrades along the continuing line, so the team-state, naming, and
reactivation mechanics it receives are the ones for its runtime rather than a
blend of both.

\subsection{Installing and upgrading}

Deployment rests on a few scripts. A bootstrap step version-checks the
environment and syncs the framework's skills and agent definitions into the
project's \texttt{.claude/}, creating or merging settings rather than
overwriting them. A thin installer wraps the first-time case; a one-button
upgrade fetches the latest version and hands off to a migration step. The
migration logic is run \emph{from the newly downloaded version}, so the code
that performs an upgrade is always the newest.

The ownership boundary from \Cref{sec:primitives} is what makes this safe. Framework-owned
paths are replaced on upgrade; user-owned paths---every workstation, the meeting
room, the archive---are never touched. Inside the one file that mixes both (the
framework README), comment markers delimit the framework's blocks---its
framework section, its work rules, and its reference material---and an upgrade
rewrites only those, leaving
project-specific content verbatim. A pre-flight
check confirms every replacement source exists \emph{before} any target is
removed, so an upgrade can never leave a project with a deleted framework and no
replacement.

\subsection{Migrations and release gates}

Upgrades that span several versions run as a chain of version-to-version steps,
each ending by writing the new version number---so a failed upgrade is
\emph{resumable}, picking up from the last step that completed. Releases
themselves are gated rather than hand-managed: a release is a dedicated commit
carrying only the version bump, the changelog entries, and the new migration
script, and a release script refuses to tag unless a set of checks all
pass---the recorded versions agree, the changelogs are non-empty, the migration
script exists and parses, the working tree is clean, and the commit being tagged
is the release commit itself. It fails closed: no
green light, no tag. These gates exist because two consecutive early releases
shipped without their migration scripts and broke the upgrade chain---exactly
the mistake the discipline now prevents.

\section{Failure Modes and Hardening}%
\label{sec:failures}

Much of this design was hardened out of real use rather than planned up front.
This section collects the failure modes that surfaced: several were described in
their own sections above, but gathered here they show the method more than any
single fix does.

\subsection{A sample of problem and mitigation}

\begin{itemize}
  \item \textbf{Teammates assumed alive after a restart.} After resuming a
    session (\texttt{/resume}), a lead would message teammates that no longer
    existed. $\to$ the liveness rule and its list of non-evidence signals
    (\Cref{sec:persistence}).
  \item \textbf{Respawns drifting to \texttt{Name-2}} \emph{(prior versions)}\textbf{.}
    A ghost entry in the persistent config collided with the new spawn.
    $\to$ clearing and rebuilding the registration in a single step before
    re-spawning.  On CC $\geq$~2.1.178 the automatic session team removes ghost
    entries on teammate exit, so this failure mode does not arise.
  \item \textbf{Work lost on an abrupt exit.} A teammate could go idle with
    unsaved progress. $\to$ the automatic idle-time checkpoint gate.
  \item \textbf{A ``timestamp didn't update'' false alarm} from a user in
    another timezone. $\to$ comparing Unix epochs instead of displayed local
    times.
  \item \textbf{Checkpoint nudge routed to the wrong workstation} when two teams
    shared a role name.  On \emph{prior versions} this was resolved by reading
    the team name from the idle event payload (T1)---a field the documentation
    now marks deprecated.  On CC $\geq$~2.1.178 the idle hook uses a three-level
    addressing chain: first the team name embedded
    in the idle event (T1), then the slug prefix of the teammate's
    \texttt{<slug>-<role>} name to derive the workstation directory
    (\texttt{\$\{name\%\%-*\}\_team}, T2), and finally a glob search with an
    exit-0 safety valve if more than one match is found (T3).
  \item \textbf{Archival buck-passing.} With several recipients, resolved
    messages were never archived. $\to$ making archival the issuer's exclusive
    duty.
  \item \textbf{A ``helpful'' lead reorganizing a teammate's files.} $\to$ the
    low-coupling rule, expanded after the incident into explicit sub-clauses.
  \item \textbf{Two consecutive releases shipped without their migration
    scripts,} breaking the upgrade chain. $\to$ the fail-closed release gates.
  \item \textbf{Rules that never reached the agents who most needed them.} The
    rule list sat in a user-owned part of each \texttt{README.md}, so an upgrade
    never refreshed it and installed copies drifted apart; teammates, whose
    context fills up fastest, had no copy on disk at all. $\to$ a second marked
    block that upgrades refresh in place, and a condensed teammate subset that
    \texttt{/spawn-team} writes into each new teammate's role file, with a
    self-heal step for workstations created before this version.
\end{itemize}

\subsection{What the catalog is, and is not}

Two things are worth stating plainly. First, this is a record of what \emph{we}
hit under sustained use, not a proof that the design is complete: the value is
the loop---run it, find a failure, fix it durably---and that loop is ongoing.
Second, the corollary: a workspace layer like this is hard to get fully right in
advance, so a method for absorbing surprises matters alongside the up-front
design. We do not claim the list is exhaustive or that the remaining failure
surface is small; \Cref{sec:limits} states the gaps we currently know about.

\section{Related Work}%
\label{sec:related}

Agent Team Work Zone draws on several active lines of work on language-model
agents. We organise them by theme and then state what this work adds.

\paragraph{Foundation models.} The agents we target are built on frontier large
language models---Claude, on which Claude Code itself runs~\citep{claudeopus48};
the GPT-5 family behind OpenAI's coding tools~\citep{gpt5}; and open-weight
families such as Llama~3~\citep{llama3}---whose breadth makes it practical to
delegate open-ended coding work to them in the first place. Our contribution is
orthogonal to the underlying model: it concerns how a \emph{team} of such models
retains state over time, and remains relevant as that capability floor keeps
rising.

\paragraph{Multi-agent LLM systems.} A line of frameworks coordinates several
language-model agents toward a shared goal. AutoGen casts applications as
conversations among customisable, tool-using agents~\citep{autogen2023}; MetaGPT
assigns role-specialised agents along a software-development assembly line driven
by standardised operating procedures~\citep{metagpt2023}; CAMEL studies
role-playing cooperation between paired agents~\citep{camel2023}; and ChatDev and
AgentVerse organise multiple agents into, respectively, a virtual software company
and a general collaboration-and-emergence
testbed~\citep{chatdev2023, agentverse2023}; and multiagent debate has several model instances propose, critique, and revise answers over rounds to improve factuality and reasoning~\citep{multiagentdebate2023}. These systems focus on \emph{how}
agents divide and coordinate work within a run; our concern is complementary and
orthogonal---making such a team, and the state it accrues, \emph{reconstructible}
after the session that hosted it ends.

\paragraph{Persisting agent state.} Closest to this work are frameworks that
already write agent state to durable storage so that work can resume. LangGraph
persists execution state in two complementary layers: a checkpointer saves a
snapshot of graph state at each super-step, scoped to a single thread, so a run
can be resumed or replayed; a store holds application-defined key--value data
outside the graph state, readable across threads, for longer-lived facts and
preferences~\citep{langgraph-persistence}. AutoGen lets a team be serialised
wholesale---calling \texttt{save\_state} on a team saves the state of all the
agents in it, the group-chat manager's message thread and turn pointers together
with each agent's model context, as a dictionary that can be written to disk and
loaded back later~\citep{autogen-state}. What these persist is execution and its
residue: a thread's state, a message history, a context window---and, in
LangGraph's store, facts meant to outlive any single thread. What none of them
persists is a unit of \emph{agency}. Our concern sits one level up---what we make
durable is not a conversation or a fact base but a \emph{teammate}: its role, its
working notes, its outstanding commitments, and its entry in a registry, so that
the participant itself, not merely the record it left behind, can be reconstituted
in a later session.

\paragraph{Reasoning, acting, and self-improvement.} Much agent capability comes
from structuring a single model's reasoning and actions, a line a recent survey reviews~\citep{reasoningsurvey2023}. Chain-of-thought
prompting elicits intermediate reasoning steps~\citep{cot2022}; self-consistency
samples and votes over multiple such chains~\citep{selfconsistency2023}; Tree of
Thoughts searches over branching reasoning states~\citep{treeofthoughts2023}; ReAct
interleaves reasoning with tool actions~\citep{react2022}; and Reflexion and
Self-Refine add verbal self-critique and iterative refinement over prior
attempts~\citep{reflexion2023, selfrefine2023}. These shape an individual agent's
loop; we instead provide the file-based substrate over which a team of such agents
persists and coordinates. A parallel line makes the reasoning itself more
controllable and economical, so agents spend tokens where they matter: hybrid
\emph{think}/\emph{no-think} models expose an explicit reasoning switch, though the
modes do not cleanly separate~\citep{hybridthinking2025} and can be disentangled
through dedicated architectural paths~\citep{pathlock2026} or steered training-free
toward an intermediate budget~\citep{midthink2026}; and the dynamics of
reinforcement-learning post-training for reasoning prove order- and
domain-sensitive~\citep{crossdomainrl2026}. These shape the reasoning substrate our
agents run on, and are orthogonal to persisting a team of such agents across
sessions. The same efficiency concern extends well beyond reasoning models, to
specialised families such as protein language models~\citep{proteinlmsurvey2026}.

\paragraph{Tool use.} A related strand teaches models to invoke external tools and
APIs: Toolformer learns when to call APIs in-line~\citep{toolformer2023}, Gorilla
targets accurate calls over large and changing API
collections~\citep{gorilla2023}, and HuggingGPT uses a controller model to
orchestrate specialised models as tools~\citep{hugginggpt2023}; and ToolLLM scales instruction tuning to thousands of real-world APIs, paired with an evaluation suite for tool use~\citep{toolllm2023}. Agent Team Work
Zone is itself realised as a thin layer of tools---hooks, scripts, and
skills---but its subject is team persistence rather than tool invocation.

\paragraph{Memory, context, and agent architecture.} Because a context window is
finite and volatile---models even use long contexts unevenly, degrading on
information placed in the middle~\citep{lostinthemiddle2023}---several systems
externalise agent state. MemGPT borrows virtual-memory paging to move information
between the context window and external storage~\citep{memgpt2023}; Generative
Agents maintain a retrieved, summarised memory stream of
experiences~\citep{generativeagents2023}; and CoALA frames language agents in terms
of modular memory and structured action spaces~\citep{coala2023}. We share the
premise that durable state must live outside the context window, but target a
different unit---an entire team's roles, working notes, and registry---and a
different failure: a session ending and taking its in-process teammates with it.
Evaluating such long-term memory is itself subtle: probed at the level of
individual facts, the dominant failure is \emph{using} retrieved evidence rather
than retrieving it~\citep{memtrace2026}.

\paragraph{Skill acquisition and reuse.} Voyager grows a library of reusable,
executable skills that an agent accumulates and later
retrieves~\citep{voyager2023}. Our layer reuses this \emph{write capability to disk}
idea at the level of process conventions and per-agent working notes, rather than
task-specific code. A recent survey maps the full lifecycle of agent skills ---
construction, composition, training integration, safety, and
evaluation~\citep{agentskillssurvey2026}.

\paragraph{Coding agents.} Autonomous software-engineering agents have advanced
quickly. SWE-bench evaluates resolving real GitHub issues~\citep{swebench2023}, and
a spectrum of systems trade autonomy against structure: SWE-agent shows the leverage
of a well-designed agent--computer interface~\citep{sweagent2024}; OpenHands offers a
general, tool-rich agent platform that also supports coordination between multiple
agents within a run~\citep{openhands2024}; CodeAct unifies an agent's
actions as executable code~\citep{codeact2024}; AutoCodeRover adds
program-structure-aware retrieval~\citep{autocoderover2024}; and Agentless shows that
a fixed, minimal pipeline can be competitive~\citep{agentless2024}---a landscape a
recent survey catalogs across 124 papers~\citep{seagentsurvey2024}, and evaluation
is itself configurable --- agent benchmarks that vary task horizon and difficulty
while cutting environment-interaction cost~\citep{agentcebench2026}. All of these
target capability \emph{within a run}, and are largely silent on continuity: what
becomes of such a team when the session hosting it ends, and how it resumes
afterwards. That continuity gap is the one this work addresses.

\paragraph{Benchmarking agents.} A parallel effort measures what such agents can
do. AgentBench evaluates language models as agents across a spread of
environments~\citep{agentbench2023}, while interactive environment suites pose
long-horizon tasks in web, general-assistant, and full-computer
settings~\citep{webarena2023, gaia2023, osworld2024}. These probe task capability
within a run, and complement the coding-specific benchmarks above.

\paragraph{What this work adds.} Broader surveys of LLM-based autonomous agents and
of multi-agent systems catalogue this design space---construction, memory, planning,
communication, tool use, and evaluation~\citep{llmagentsurvey2023, multiagentsurvey2024}.
Against that backdrop, Agent Team Work Zone
is not another agent architecture but a \emph{persistence and management layer}: it
takes an existing Agent Teams primitive and is designed to make a team durable,
recoverable, and auditable across sessions through files, checkpoints, and a
one-command reactivation flow (\Cref{sec:persistence}); we describe those
mechanisms rather than measure their reliability (\Cref{sec:limits}). A broader
map of the LLM and agent-harness landscape these systems inhabit is given in
\Cref{app:background}.

\section{Limitations and Future Work}%
\label{sec:limits}

The design has edges we know about; stating them is part of the same posture as
the rest of the manual.

Several limitations are deliberate trade-offs. The idle checkpoint gate fails
open when a teammate has no checkpoint file yet, so a brand-new teammate that
has never checkpointed gets no automatic nudge---a conscious choice (never block
a teammate out of doubt) that leaves a gap for the freshly-onboarded. And the
contract that keeps the checkpoint \emph{writer} and
\emph{reader} in agreement is a documentation convention, not a checked one: a
developer who changes one side and forgets the other would silently weaken
recovery, and a consistency test would help.

Others are open questions. Two internal detections lean on text patterns rather
than structured state: whether a document's completion checklist is fully
checked, and whether the installer's sections are already present in a project's
\texttt{CLAUDE.md}. Two earlier entries have since closed themselves, both from the prior API, whose
reactivation began by clearing and rebuilding the team registration. In the
first, the clearing call tolerated a no-op, since a freshly restarted lead
usually has no team context to clear---the normal path, not a degradation---while
a failed rebuild aborted the run. In the second, the lead had to infer whether a
named reactivation should skip that clearing step, because registration could
not be queried. Version~2.1.178 removed both tools and the session-scoped team
removed the step, so neither question arises on the current line. None of
these are hidden; they are the current entries in the same find-and-fix loop the
rest of the design came from.

Most importantly, this manual is a design, not an evaluation. We have described
mechanisms and the reasoning behind them and illustrated them through how the
layer is used, but we have not measured how much they help, compared them
against alternatives, or studied how people work with them. Empirical
evaluation---recovery reliability under real interruptions, the overhead the
discipline adds, and usability of human-in-the-loop operation---is the main
direction for future work, alongside a tighter lead/teammate protocol, less
prompt-writing overhead, and keeping the framework simple enough that agents
reliably follow it. Established methodology for evaluating language models offers
a starting point~\citep{llmevalsurvey2024}, though it targets task quality rather
than the recovery and continuity properties at issue here. A more autonomous mode, with the human further out of the
loop, is left for when models are capable enough to be trusted with it.

\appendix
\renewcommand{\thesection}{\Alph{section}}%
\renewcommand{\thesubsection}{\Alph{section}.\arabic{subsection}}%
\setcounter{section}{0}%
\section{Background on LLMs and Agent Harnesses}%
\label{app:background}

The Related Work section (\Cref{sec:related}) situates this layer among systems at
its own level---multi-agent frameworks, reasoning methods, coding agents. This appendix
steps back to sketch the two substrates the layer rests on: the model capability
that makes it feasible to delegate open-ended coding work at all
(\Cref{app:llms}), and the harness scaffolding that turns a model into an agent,
and a set of agents into a team (\Cref{app:harness}). Agent Team Work Zone adds a
persistence layer above both, and is complementary to the advances mapped here.

\subsection{Large Language Models}%
\label{app:llms}

The agents this layer coordinates are built on large language models whose
capability rises with scale. Scaling laws established that loss falls predictably
with model size, data, and compute~\citep{scalinglaws2020}; a compute-optimal
analysis rebalanced parameters against training tokens~\citep{chinchilla2022};
and some capabilities appear only past a scale threshold rather than
improving smoothly~\citep{emergent2022}---the rising ``capability floor'' the
main text refers to.

Raw next-token prediction became usable for delegation through two shifts.
In-context learning let a model perform a task from a few examples in the prompt,
with no weight updates~\citep{gpt3-2020}; a survey traces how this ability works and how it has been understood since~\citep{iclsurvey2024}. Instruction tuning---finetuning on tasks
phrased as natural-language instructions---then made models follow unseen
instructions zero-shot~\citep{flan2021}, and reinforcement learning from human
feedback~\citep{rlhf2017} aligned outputs with human intent, yielding
instruction-following models that act on a natural-language task
description~\citep{instructgpt2022}; later work simplified this alignment step by optimizing directly on preference pairs, dispensing with a separate reward model~\citep{dpo2023}. This is the capability the layer assumes when
a lead delegates in natural language.

Efficiency is itself a live concern for agents, since a team's cost and latency
scale with model size and how long each teammate thinks. A broad literature makes
language-model inference cheaper---key--value-cache paging and IO-aware attention
raise throughput~\citep{vllm2023, flashattention2022}, post-training quantization
cuts the compute and memory of a trained model~\citep{gptq, proteinlmsurvey2026},
and sparse mixture-of-experts architectures buy capacity below dense
cost~\citep{moe-switch2021}---and the layer responds by matching each teammate's
model to its work (\Cref{sec:roles}).

Two broader treatments give the full picture: a foundation-models framing of the
shift to large, adaptable pretrained models~\citep{foundationmodels2021}, and a
comprehensive survey of large language models~\citep{llmsurvey2023}.

\subsection{Agent Harnesses}%
\label{app:harness}

Between a language model and a working agent sits a \emph{harness}: the
scaffolding that turns single-shot text generation into a loop that reasons,
calls tools, and carries state. A survey of emerging agent architectures maps the
choices this layer makes---single- versus multi-agent structure, leadership,
communication, and the planning, execution, and reflection
phases~\citep{agentarchsurvey2024}. Agent Team Work Zone sits one level above such
a harness: it does not implement the loop, but persists the team the loop
produces.

One harness capability is \emph{tool use}---letting a model act by calling
external functions and APIs. A survey frames tool learning as a general
capability~\citep{toollearningsurvey2023}; the Related Work section cites specific
methods. Another is \emph{memory}---giving an agent state that outlives a single
context window. A survey catalogs memory mechanisms for LLM
agents~\citep{agentmemsurvey2024}; the Related Work section contrasts this
layer's team-level state with those agent-level mechanisms.

The reasoning loops, cognitive-architecture framings, and agent-computer
interfaces that make up the rest of a harness are covered in the Related Work
section. Across all of them the harness is what \emph{runs} an agent; this layer
adds what \emph{carries} a team of them past the session that ran it.

\printbibliography[title={References}]

\end{document}